\documentclass[letterpaper,10pt,conference]{ieeeconf}

\IEEEoverridecommandlockouts
\usepackage{amsmath,amssymb,bm}
\usepackage{graphicx}
\usepackage{booktabs}
\usepackage{array}
\usepackage{tabularx}
\usepackage{xcolor}
\usepackage{pifont}
\usepackage{microtype}
\usepackage{cite}
\usepackage{multirow}
\usepackage{textcomp}
\usepackage{siunitx}
\usepackage{svg}

\graphicspath{{figures/}}

\title{\LARGE \bf
MicroHookACT: Monocular Microscopic Vision Guided Visuomotor Policy for Flexible Microelectrode Hooking}

\author{Yitong Chen$^{1,2}$, Fangbo Qin$^{*1,2,3}$, Yang Wang$^{4}$, Ruihua Hu$^{1}$, Kui Zhang$^{1}$, Shan Yu$^{1,2,3}$% <-this % stops a space
\thanks{$^{1}$Institute of Automation, Chinese Academy of Sciences, Beijing 100190, China. 
$^{2}$School of Artificial Intelligence, University of Chinese Academy of Sciences, Beijing 100049, China. 
$^{3}$State Key Laboratory of Brain Cognition and Brain-Inspired Intelligence Technology, Shanghai 200031, China.
$^{4}$Institute of Semiconductors, Chinese Academy
of Sciences, Beijing 100083, China.} }

\centerfigcaptionsfalse
\begin{document}
\bstctlcite{BSTcontrol}
\clubpenalty=0          % 允许段落首行单独留在上一页/栏底部
\widowpenalty=0         % 允许段落末行单独出现在下一页/栏顶部
\displaywidowpenalty=0  % 取消陈列公式前的特殊孤行惩罚
\maketitle
\thispagestyle{empty}
\pagestyle{empty}
% \raggedbottom
\begin{abstract}
Automated needle-loop hooking is a critical step in flexible microelectrode (FME) implantation. This paper presents MicroHookACT, an imitation learning-based visuomotor policy for automated 3D hooking under monocular microscopic vision. First, a unidirectional hooking strategy exploits defocus cues and optical-axis guidance to enable palpation-free precise alignment and contact-rich threading. Second, an action-supervised object attention module built on a frozen ViT backbone learns to focus on the micro-needle tip and micro-loop directly from human demonstrations, without requiring manual visual annotations for training. Third, attention-centered global coarse and local fine features are dynamically weighted according to predicted action progress, enabling a single ACT policy to adapt to changing defocus blur and visual requirements throughout the operation. In the experiments, visuomotor policies were trained on 60 human demonstrations and evaluated under five setups with varying difficulties. Our MicroHookACT framework achieved the highest overall success rate of 96.7\% with an average execution time of 11.5 s. These results demonstrate the potential of visuomotor policy learning for micron-level control under varying operating conditions.

\end{abstract}

%%%%%%%%%%%%%%%%%%%%%%%%%%%%%%%%%%%%%%%%%%%%%%%%%%%%%%%%%%%%%%%%%%%%%%%%%%%%%%%%
\section{INTRODUCTION}

A flexible microelectrode (FME) is a thread-like neural probe that is more flexible than a human hair and connects the cerebral cortex to electronic chips. With promising biocompatibility and long-term stability, FMEs have been increasingly used in invasive brain–computer interfaces \cite{liu2024flexible}. To implant FMEs with extremely low Young’s moduli into the cerebral cortex, the “sewing machine” implantation paradigm was proposed. In this paradigm, a rigid micro-needle hooks the loop structure at the distal end of an FME and carries it into the cerebral cortex. The micro-needle is then withdrawn independently, leaving the FME implanted within the cortex \cite{hanson2019,musk2019,chen2025delicate, chen2025anomaly}. Needle-loop hooking is a critical yet challenging step because the micro-loop at the FME tip has an inner diameter of only approximately \SI{50}{\micro\meter} and is highly susceptible to deformation. Meanwhile, control of the implantation needle must accommodate both micromanipulation accuracy and the centimeter-scale workspace.

A series of micromanipulation methods based on microscopic vision guidance have been proposed for the automated FME hooking task. Most methods first localize image features of the FME and micro-needle, and then perform microscopic visual servoing based on these features. An et al. \cite{an2024microscopic} developed a monocular microscopic visual servoing system for floating electrode assembly, using image processing to localize the needle tip and training a target detection network on a large number of annotated images to detect the electrode hole. Qin et al.\cite{qin2023automated} proposed a visual framework based on few-shot learning that generated keypoint and contour prototypes from annotated support images. A two-stage binocular microscopic visual servoing strategy enabled fully automated hooking through coarse alignment at low magnification followed by fine alignment at high magnification. Although this framework reduced the annotation requirements, it is sensitive to distractors and occlusion. Han et al.\cite{han2025intelligent} developed a stereo microscopic vision system for hooking multiple FMEs, combining few-shot instance segmentation with synthetic image augmentation to extract keypoints for visual guidance. In a subsequent study, Han et al.\cite{han2026closedloop} adopted the complete FME as the segmentation target and synthesized numerous occluded images from a small set of real training images to improve localization under occlusion. However, the real samples still required instance-level annotations which imposed more annotation burden.

\begin{figure}[t]
    \centering
    \begin{minipage}{0.49\textwidth}

    \includegraphics[width=\columnwidth]{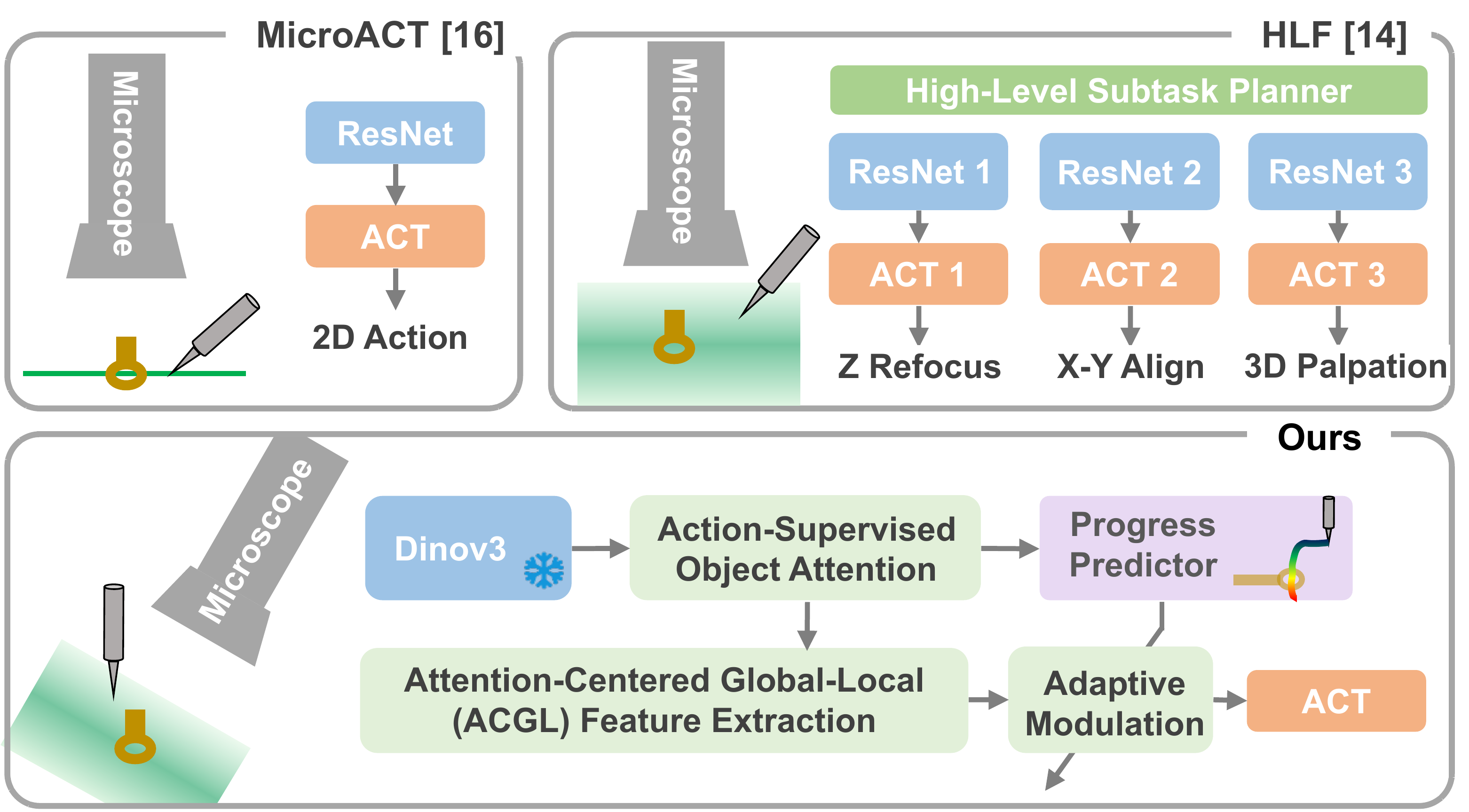}
    \vspace{-16pt}
    \caption{Monocular microscopic vision guided visuomotor policies. The green gradient region indicates the microscope’s depth of field, with darker shades representing positions closer to the focal plane. }
    \label{fig:intro}
    \end{minipage}
\end{figure}

In the aforementioned methods, intelligent visual perception and automated control strategy are developed separately and subsequently integrated, resulting in the following limitations: 1) the visual perception module requires image collection, annotations, model training and parameter tuning, while the automated control system also requires design and parameter tuning, making the overall deployment labor-intensive and time-consuming; 2) the control system relies on the results of explicit keypoint detection or segmentation mask extraction, which can fail in the presence of defocus blur and mutual occlusion between the needle and FME.

Imitation learning-based visuomotor policies provide an important approach to vision-guided manipulation by establishing a direct mapping from visual observations to actions. They eliminate the need for dedicated dataset construction and training for the visual module by automatically learning implicit features that guide actions directly from human demonstration data, while jointly training an action policy tightly coupled with visual perception \cite{zare2024survey}. Action Chunking Transformer (ACT) is a widely adopted visuomotor policy. Based on a Transformer architecture, it predicts a sequence of future continuous actions conditioned on images and robot proprioceptive states, and uses a conditional variational autoencoder (CVAE) to model multimodal action distributions \cite{zhao2023act,fu2024mobilealoha,buamanee2024biact}. 

Luo et al. \cite{luo2026hierarchical} first introduced the ACT policy to the monocular microscope-guided FME hooking action learning, and proposed a hierarchical learning framework that decomposed the complete hooking operation into three high-level subtasks: refocusing, alignment, and palpation. A high-level planner selected next subtask, while the low-level execution module assigned an independent ACT policy network to each subtask. Each network predicted actions in the camera coordinate frame to enable precise control within its corresponding stage. However, this method required manual divsion of demonstration data into stages and the training of multiple independent ACT policies. Moreover, the high-level planner required continued refinement through a human-in-the-loop approach, increasing the manual effort required for practical deployment. In addition, imitation learning has been increasingly applied to other micromanipulation tasks, including gaze guidance \cite{an2024skill}, 2D pose adjustment \cite{long2025microact}, oocyte rotation \cite{mori2024realtime}, long-horizon cell membrane stripping \cite{zhang2025sail}, and pseudo oocyte gripping \cite{eljuri2025haptic}.

\subsection{Hooking Strategy under Monocular Microscope}
\begin{figure*}[t]
    \centering
    \centering
    \includegraphics[width=\linewidth]{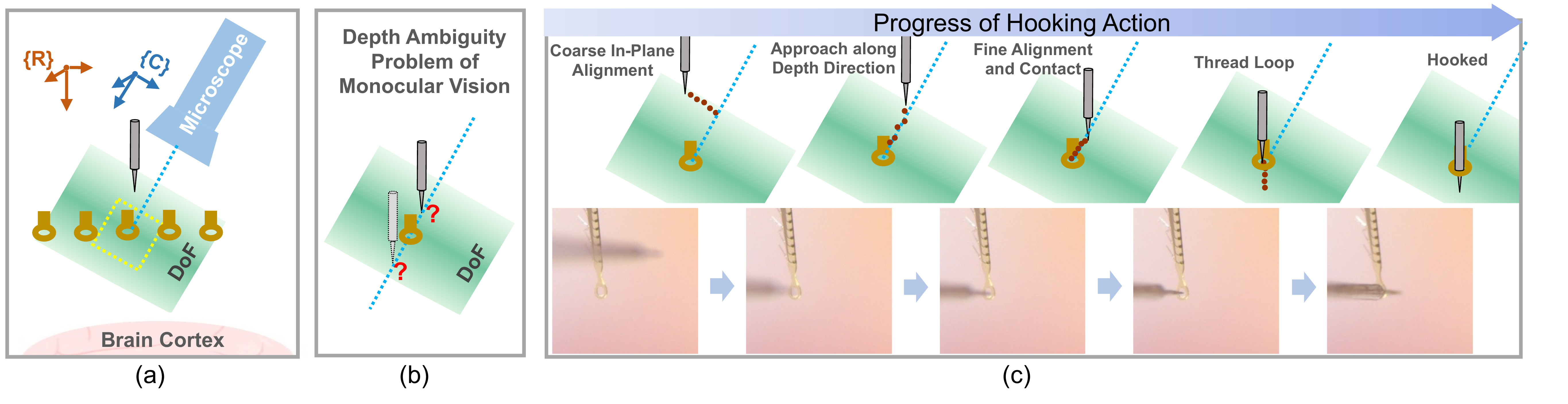}
    \vspace{-25pt}
    \caption{FME implantation setup and unidirectional hooking strategy. (a) Spatial relationship between micro-needle, FMEs, microscope, and brain cortex. The region of interest (ROI) view and optical axis of the camera are depicted with a dashed box and line, respectively. The green gradient region indicates the microscope’s depth of field (DoF), with darker shades representing positions closer to the focal plane. (b) A monocular view cannot distinguish which of the two micro-objects is closer to the camera. (c) The unidirectional needle-loop hooking strategy is divided into four stages, which are modeled with a single visuomotor policy. }
    \label{fig:progress}
\end{figure*}

Following \cite{luo2026hierarchical,long2025microact}, we are primarily motivated to address three issues:
1) While \cite{luo2026hierarchical} employed a palpation subtask to achieve 3D hooking despite the depth ambiguity of monocular microscopic vision, we aim to achieve unidirectional hooking without trial-and-error motions, thereby improving operational efficiency.
2) We aim to enable the visual module to automatically learn attention to micro-objects from demonstration data, focusing visual encoding on these objects rather than encoding only the full image frame.
3) As defocus blur and inter-object distance vary throughout the operation, the visual information relevant to action generation also shifts. We aim to accommodate these changes adaptively within a single model, without training separate models for different stages as in \cite{luo2026hierarchical}. Our main contributions are as follows:

1) We design a unidirectional hooking strategy that exploits the gradient of the microscope’s focus field and the directional guidance provided by its optical axis, enabling palpation-free precise alignment and contact-rich threading under monocular microscopic guidance.

2) Building on a pretrained ViT visual backbone, we propose action-supervised object attention, which automatically learns spatial attention to critical object parts during imitation learning through multiple iterations and global image aggregation. This mechanism reduces the influence of environmental disturbances and changes in operating conditions on action generation.

3) We propose extracting global coarse and local fine features from the attended features for action conditioning and dynamically modulating their relative weights using predicted action progress. This enables a single model to adapt to variations in defocus blur and shifts in relevant visual information throughout the hooking process.

%%%%%%%%%%%%%%%%%%%%%%%%%%%%%%%%%%%%%%%%%%%%%%%%%%%%%%%%%%%%%%%%%%%%%%%%%%%%%%%%
\section{METHODOLOGY}

As shown in Fig.~\ref{fig:progress}(a), flexible microelectrode (FME) threads are arranged above the brain cortex. Each FME has a micro-loop structure at its end. To implant the soft and slender FME into the brain cortex, a micro-needle tip is used to hook the micro-loop and carry the FME thread into the tissue. We use a monocular microscopic vision system to perform 3D microneedle–micro-loop hooking. Compared with stereo microscopic vision, monocular microscopic vision simplifies setup and calibration. However, it lacks depth information and suffers from the \textit{depth ambiguity} problem: the relative depth between the microneedle tip and the micro-loop cannot be directly inferred from a single image, as illustrated in Fig.~\ref{fig:progress}(b). However, the shallow depth of field (DoF) of the microscopic camera~\cite{yang2025fine} provides depth cues, making 3D micromanipulation under a monocular microscope feasible.

Unlike the palpation-based approach~\cite{luo2026hierarchical}, which relies on trial-and-error, we propose a unidirectional hooking strategy that avoids such iterative adjustment. The procedure is illustrated in Fig.~\ref{fig:progress}(c). First, the microscope is focused on the micro-loop. The microneedle tip is positioned approximately \qtyrange{500}{2000}{\micro\meter} above the target FME. Although the initial relative position between the microneedle and the micro-loop is uncertain, the fact that the micro-loop is located deeper than the micro-needle tip serves as a known prior.

In the first stage, coarse in-plane alignment is achieved by moving the microneedle in the XY plane of the camera coordinate frame $\{C\}$, thereby approximately aligning the defocused, blurred tip with the center of the in-focus micro-loop. In the second stage, the microneedle approaches the micro-loop along the depth direction, i.e., the Z direction of $\{C\}$, while maintaining in-plane alignment. During this approach, the tip moves along an approximately straight line that passes through the micro-loop center and is parallel to the optical axis, and its image becomes progressively sharper. In the third stage, both the tip and the micro-loop are clearly imaged, enabling fine alignment. The tip continues to advance in small steps along the depth direction until it comes into contact with the inner edge of the micro-loop. In the fourth stage, the microneedle moves along the Z direction of the robot coordinate frame $\{R\}$, while making small adjustments along the X direction of $\{R\}$, enabling the microneedle tip to smoothly thread through the micro-loop.

Throughout this process, the microneedle tip moves unidirectionally along the depth direction. The coarse-to-fine alignment ensures that the tip reaches the micro-loop center and continues to pass through the loop, eliminating the need for palpation-based back-and-forth adjustment. However, this strategy requires high positioning accuracy and visual guidance even when the tip is heavily defocused and blurred.

\subsection{MicroHookACT Framework}

Following~\cite{luo2026hierarchical} and~\cite{long2025microact}, we adopt an ACT policy conditioned on a single microscopy image, where the output action chunk represents a motion trajectory in the microscopy camera coordinate frame $\{C\}$. Since microneedle-micro-loop hooking is performed through relative motion, the action chunk is defined as a relative motion trajectory of the micro-needle in $\{C\}$ rather than an absolute one. Furthermore, relative motion does not require the robot's states as input. Accordingly, MicroHookACT is formulated as
\begin{equation}
a_{t:t+k}=\pi_{\mathrm{ACT}}\left(v_t,z_t\right),
\label{eq:microhookact}
\end{equation}
where $v_t$ is the visual representation obtained from a single microscopy image, and $z_t\sim\mathcal{N}(0,1)$ is a latent variable. The action chunk $a_{t:t+k}$ is a continuous 3D motion trajectory whose origin is the current needle-tip position.

\begin{figure*}[t]
    \centering
    \begin{minipage}[c]{\textwidth}
        \centering
    \includegraphics[width=\linewidth]{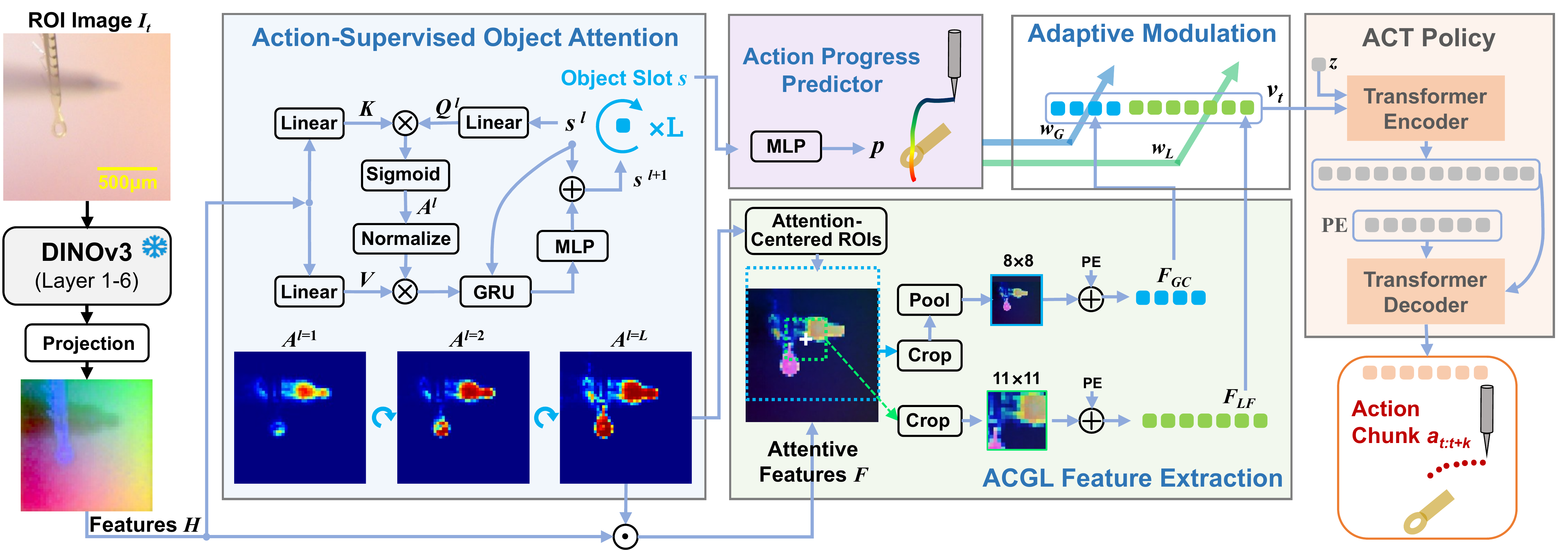}
    \end{minipage}\hfill
    \caption{MicroHookACT framework.}
    \label{fig:framework}
\end{figure*}

The raw feature map produced by a CNN or ViT contains numerous pixels that are irrelevant to action guidance. Conditioning the policy on all these features results in a high-dimensional input, which can degrade policy generalization and increase computational cost. As shown in Fig.~\ref{fig:framework}, we use a frozen DINOv3 ViT-B~\cite{simeoni2025dinov3} to extract a transferable pretrained ViT feature map from the input microscopy image. We then apply a $1\times1$ convolutional layer to project the features from 768 to $D=256$ dimensions, yielding the projected feature map $H$. Next, we introduce an action-supervised object attention module to focus the model's attention on the microneedle tip and micro-loop while suppressing features from the background and irrelevant parts, producing the attentive feature map $F$.

Based on the center of the object attention distribution, we extract global and local features from the attentive feature map $F$. The global features guide the policy during coarse alignment and depth approach when the global positions of the microneedle tip and micro-loop are unknown and substantial defocus blur is present. The local features support fine alignment and needle threading at high spatial resolution when the micro-needle tip is already close to the micro-loop. Furthermore, to adaptively leverage these two types of features, we do not merely concatenate them. Instead, we employ the object query token to predict the progress of the current hooking action and dynamically adjust the importance of the global and local features based on this progress. This produces the modulated visual features $v_t$, which serve as the conditioning input to the ACT policy.

MicroHookACT is trained end-to-end using only human teleoperation data as supervision, without requiring any manual visual annotations. During training, the DINOv3 backbone remains frozen to preserve generalizable visual priors. The teleoperation commands include incremental alignment motions in X, Y, and Z in the camera coordinate frame $\{C\}$ and threading motions along the Z axis in the robot frame $\{R\}$. When the operator observes that the microneedle tip has contacted the inner rim of the micro-loop after fine alignment, the commands switch from alignment motions in $\{C\}$ to threading commands for the first time. The moment of this first switch is used as the indicator of fine-alignment completion and can subsequently serve as a supervision signal for task progress estimation.

\subsection{Action-Supervised Object Attention}

For microneedle-micro-loop hooking, effective visual cues come from the microneedle tip and micro-loop, which together occupy only a small fraction of the image ($\sim10\%$). To focus the visual conditioning precisely on the target micro objects, we propose an action-supervised object attention mechanism inspired by the iterative object querying and feature aggregation approach of Slot Attention~\cite{locatello2020slot}. Specifically, we employ a learnable object slot $s \in \mathbb{R}^{D}$ as a query relevant to the action policy, and use this slot to aggregate visual information from $H$. Unlike the original Slot Attention, our mechanism is learned during imitation learning using only the action loss, without any image-level supervision, and can automatically learn to identify key spatial regions relevant to the action policy. In preliminary experiments, we observed that when actions were the only supervision signal, using multiple object slots tended to prevent different slots from consistently specializing to distinct micro-objects. We therefore use a single object slot to attend to all action-relevant locations. This removes the need for a softmax-based slots competition mechanism and allows to directly use a sigmoid function to compute a single-channel spatial attention map.

As shown in Fig.~\ref{fig:framework}, we flatten the feature map $H$ into a sequence of $N$ feature vectors and apply separate linear projections to obtain $K$ and $V$, both of dimension $D$. Meanwhile, we project $s^l$ into $Q^l$ through a linear projection. We then compute the object attention map using sigmoid function, as follows,
\begin{equation}
A^l=\operatorname{sigmoid}\!\left(
\frac{Q^lK^T}{\sqrt{D}}
\right).
\label{eq:attention}
\end{equation}

To aggregate the value features, we first normalize $A^l$ along the spatial dimension and then multiply the normalized attention weights by  $V$, yielding an update feature:
\begin{equation}
u^l=\frac{A^lV}{\sum_{i=1}^{N}A_i^l}\in \mathbb{R}^{D}.
\label{eq:update}
\end{equation}
We then use a gated recurrent unit (GRU)~\cite{cho2014learning} and a residual MLP to update $s^l$, yielding the object slot at iteration $l+1$:
\begin{equation}
s^{l+1}=s^l+\operatorname{MLP}\left(\operatorname{GRU}(s^l,u^l)\right).
\label{eq:query-update}
\end{equation}
This process is repeated for $L=3$ iterations to obtain action-relevant spatial attention. Through these iterative updates, the object slot also aggregates action-relevant visual context from the current scene, which is subsequently used for action progress prediction.

\subsection{Attention-Centered Global-Local Feature Extraction}

We compute the element-wise product of the object attention map $A^L$ and the feature map $H$ to suppress features in irrelevant regions, yielding attentive features $F=H\odot A^{l=L}$. To reduce the number of features fed into the downstream action policy while maintaining effective representations of both global relative positions and fine local interactions, we propose the ACGL feature extraction method. This method is inspired by the visual perception patterns of human operators during teleoperation: in the early stages of the action, operators primarily focus on the approximate global needle-loop relative positions, whereas in the later stages, they focus on small local offsets and fine interactions around the micro-loop. In addition, the visual region of interest (ROI) also depends on the spatial distribution of the objects. For example, when both the micro-needle and micro-loop are located in the upper half of the image, the operator's visual ROI also lies in the upper half, even if the two objects remain far apart. Therefore, we first compute the centroid $(x_c,y_c)$ of the object attention distribution in $A^L$ and use this centroid to obtain the global and local ROIs.

To obtain a global coarse representation, we crop a global ROI centered at $(x_c,y_c)$ that has the same spatial dimensions as $F$ and apply zero padding to beyond-boundary region. We then perform average pooling along to reduce the spatial dimensions to $8\times8$ feature map. After adding a 2D position embedding (PE) and flattening the result, we obtain 64 global coarse features $F_{GC}$, which describe the overall relative layout over a large spatial extent. Aligning the ROI with the attention centroid provides a relative position representation that is less sensitive to changes in absolute position.

To obtain a local fine representation, we crop a $11\times11$ local ROI centered at $(x_c,y_c)$ while preserving the original spatial resolution, zero-padding any region that extends beyond the boundaries of ${F}$. This ROI can encompass both the micro-needle tip and micro-loop in the later stages of the action. After adding a 2D PE to the features within the local ROI and applying a flatten operation, we obtain 121 local fine features $F_{LF}$, which primarily capture the relative relationship between the micro-needle and the micro-loop at high resolution when they are close to each other.

\subsection{Action Progress-Driven Adaptive Modulation}

Needle-loop hooking consists of distinct stages with different motion patterns. Although the overall motion trajectory is relatively simple, adaptive visual guidance at different stages is essential for reliable and precise micro-needle motion. Therefore, before concatenating $F_{GC}$ and $F_{LF}$ into the visual observation and feeding it into ACT, we propose to adaptively modulate $F_{GC}$ and $F_{LF}$ according to the action progress. The aim is to smoothly shift the observation focus from $F_{GC}$ to $F_{LF}$ as the action proceeds, enabling a single model to operate consistently across different stages.

For action progress prediction, we use an MLP to estimate the progress $p$ from the object slot vector $s$, which aggregates action-relevant visual information from the current scene. Learning action progress prediction requires auxiliary supervision, but we aim to avoid any additional data annotation burden beyond teleoperation. During human teleoperation, the time when the first threading motion command is issued can be recorded as the time $t_{thr}$ at which alignment between the micro-needle tip and micro-loop center is completed. Since the needle tip is aligned with the micro-loop center at this time, its absolute position
 in $\{R\}$ , $P_{thr}=(x_{thr},y_{thr},z_{thr})$, can be used to represent the actual position of the micro-loop. The distance between $P_{thr}$ and the absolute position at any other time $t$, $P_t=(x_t,y_t,z_t)$, indicates the action progress. Accordingly, the ground-truth action progress in the teleoperation data is computed as
\begin{equation}
\bar{p}_t=\frac{1}{\rho}\operatorname{sgn}(t-t_{thr})\cdot
\left\|P_t-P_{thr}\right\|_2,
\label{eq:progress}
\end{equation}
where $\rho$ is a scaling factor whose default value is the needle-tip length, i.e., \SI{300}{\micro\meter}. Here, $\operatorname{sgn}$ is the sign function, which indicates whether the current action progress is before or after $t_{thr}$.

During deployment, $p$ is used to compute the modulation coefficients for the ACGL features online:
\begin{equation}
w_{L}=\operatorname{sigmoid}\!\left(\frac{p-\delta_p}{\tau_p}\right),
\qquad w_{G}=1-w_{L},
\label{eq:routing}
\end{equation}
where $\delta_p$ and $\tau_p$ are the progress bias and temperature coefficient, respectively, with $\tau_p=0.2$. The default value of $\delta_p$ is $-1.5$, because the needle tip gradually comes into focus when the needle-loop distance is approximately $1.5$ times the needle-tip length $\rho$. Finally, we use these weights to modulate the ACGL features and concatenate the results to obtain the visual observation representation $v_t$:
\begin{equation}
v_t=\operatorname{concat}
\left(w_{G}\times F_{GC},w_{L}\times F_{LF}\right)
.
\label{eq:mgl}
\end{equation}

\subsection{Imitation Learning of MicroHookACT}

We integrate the above visual model with ACT~\cite{zhao2023act} to form a visuomotor policy for imitation learning of needle-loop hooking. We customize the architectural parameters of original ACT, using 3 encoder layers and 2 decoder layers. The hidden dimension and feedforward dimension are set to 256 and 1024, respectively. The predicted action chunk size $k$ is set to 10. The actually executed steps are 2. Using the least-squares method, we calibrate the rotation transformation matrix ${}^{C}\mathcal{R}_{R}$ from the robot coordinate frame $\{R\}$ to the camera coordinate frame $\{C\}$ to transform relative motions between the two coordinate frames. Each human teleoperation demonstration consists of a sequence of image frames, absolute 3D positions in $\{R\}$, and threading flag values, denoted as $\mathcal{D}=\{I_t,P_t,b_{thr}\}_{t=1:T}$. From these data, we generate the ground-truth action chunk:
\begin{equation}
\bar{a}_{t:t+k}={}^{C}\mathcal{R}_{R}
\{
P_{t+i}-P_t
\}_{i=1:k}.
\label{eq:act_gt}
\end{equation}

Input frames paired with their corresponding ground-truth action chunks are used for imitation learning. The loss function comprises an $L1$ loss $\mathcal{L}_{\mathrm{action}}$ for action chunk reconstruction, a $KL$ divergence loss $\mathcal{L}_{\mathrm{CVAE}}$ for regularizing the CVAE encoder, and an $L2$ loss $\mathcal{L}_{\mathrm{progress}}$ for action progress prediction. The total loss is a weighted sum of these three terms:
\begin{equation}
\label{eq:loss}
\mathcal{L}=\mathcal{L}_{\mathrm{action}}
+\beta\mathcal{L}_{\mathrm{CVAE}}
+\gamma\mathcal{L}_{\mathrm{progress}}.
\end{equation}

During training, data augmentation includes random color jitter, illumination variations, Gaussian blur, and small changes in scale and translation. The trained policy outputs relative motion trajectories in $\{C\}$, which we transform back to $\{R\}$ for execution. In on our preliminary study, temporal ensemble did not provide a clear benefit for this task, and we therefore didn't use it.

%%%%%%%%%%%%%%%%%%%%%%%%%%%%%%%%%%%%%%%%%%%%%%%%%%%%%%%%%%%%%%%%%%%%%%%%%%%%%%%%
\section{EXPERIMENTS AND RESULTS}
\begin{figure}[t]
    \centering
    \begin{minipage}{0.49\textwidth}
    
    \includegraphics[width=\columnwidth]{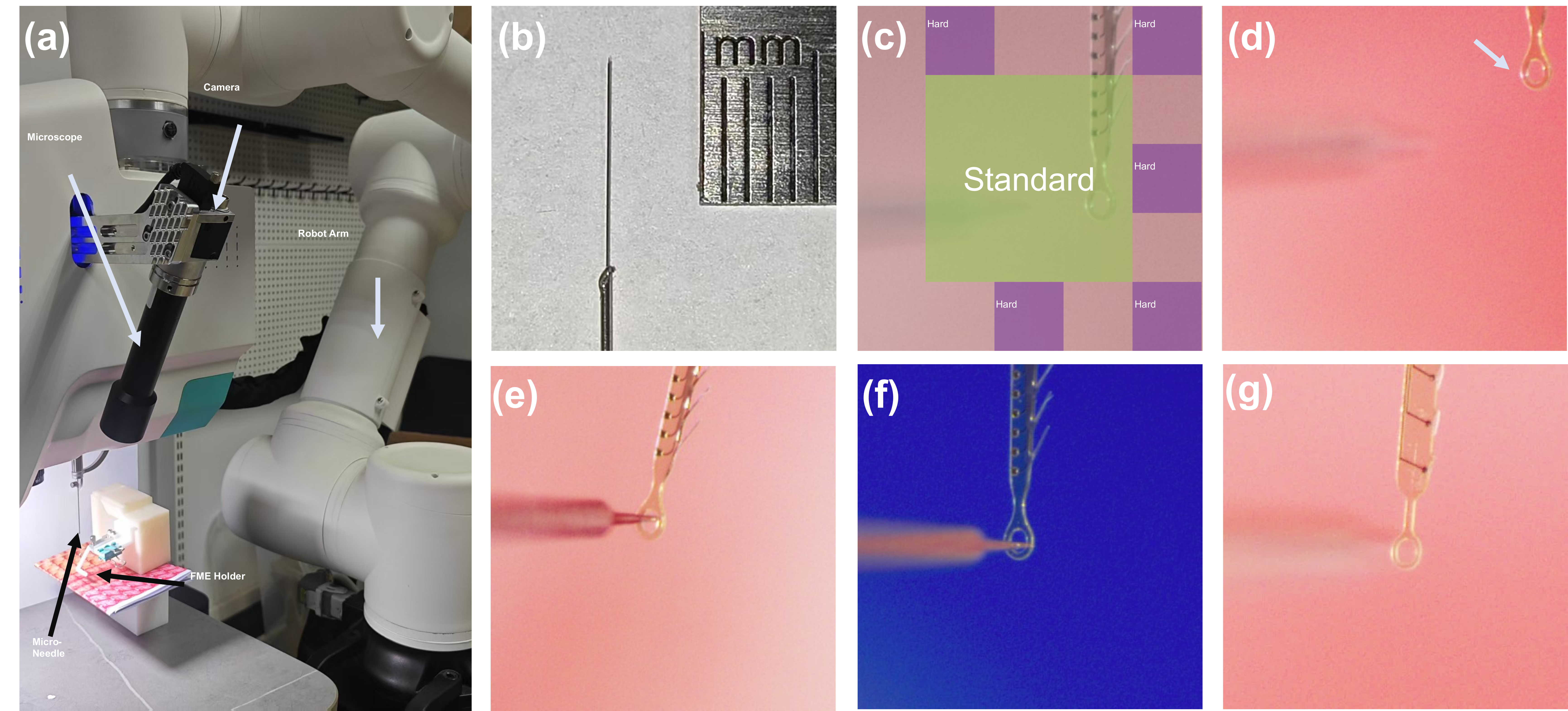}
    \vspace{-16pt}

    \caption{Experiment platform and evaluation setups.}
    \label{fig:platform}
    \end{minipage}
\end{figure}

\subsection{Experiment Platform}
As shown in Fig.\ref{fig:platform}(a), the experimental platform was a microscopic surgical robot designed for animal experiments in neuroscience research. The camera was a Basler acA2440-20gc with a pixel size of $3.45  \times$\SI{3.45}{\micro\meter}. The lens was a telecentric microscope KW-T1.5X110 with a magnification of $1.5\times$ and a depth of field of \SI{420}{\micro\meter}. As shown in Fig.\ref{fig:platform}(b), the micro-needle had a diameter of \SI{100}{\micro\meter} and a stepped tip with \SI{300}{\micro\meter} length and \textless\SI{30} {\micro\meter} diameter at its distal end. The Micro-needle was driven by Kohzu XA04A precision stages to achieve translation in three degrees of freedom with a motion resolution of \SI{1}{\micro\meter}. A robotic arm was used to adjust the position and orientation of the microscopic surgery head relative to the FME and cranial window. The FME was bonded to a cantilever support to hold its micro-loop in a fixed position. The Z-axis of the robot coordinate frame $\{R\}$ pointed vertically downward, while the Z-axis of the camera coordinate frame $\{C\}$ was tilted downward at an angle of approximately $30^\circ$ relative to the Z-axis of $\{R\}$. The host computer had an NVIDIA RTX 3070 GPU for the visuomotor policy inference.

\begin{figure*}[t]
    \centering
    \includegraphics[width=\linewidth]{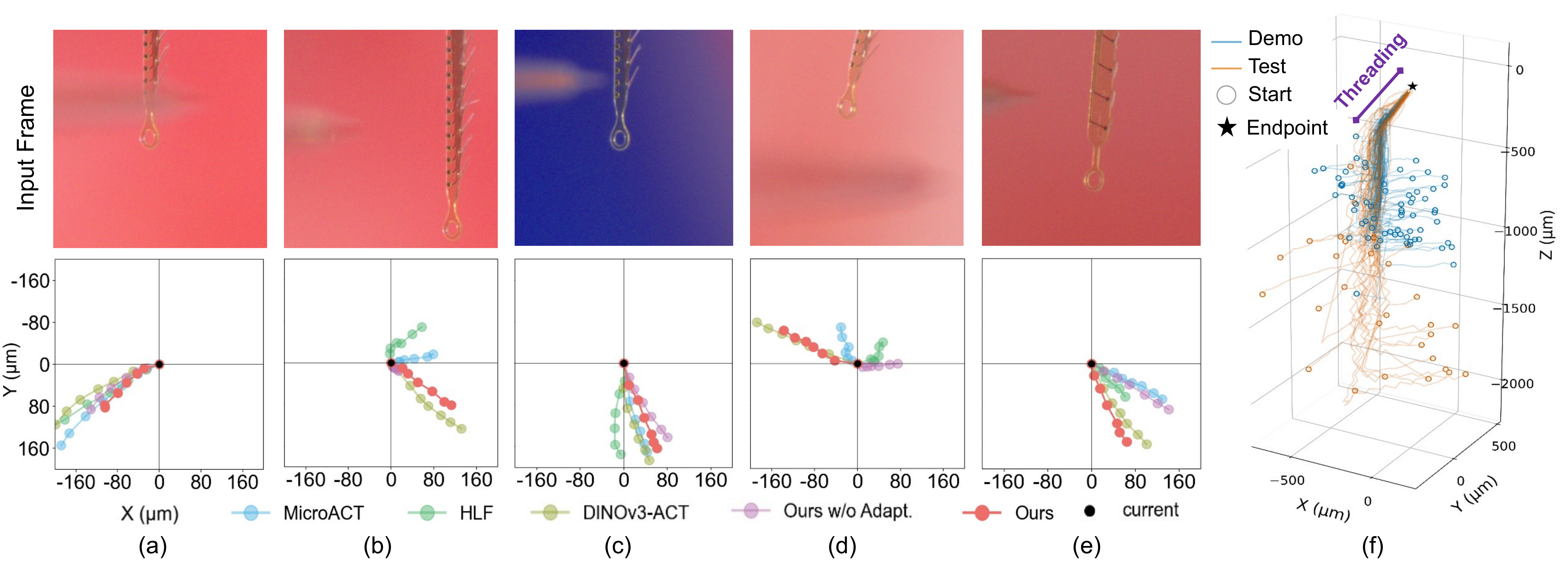}
    \vspace{-25pt}
    \caption{Action trajectory visualization. (a-e) Relative motion trajectories in \(\{C\}\) conditioned on the input frame. The movement direction reflects the correctness of the coarse approaching actions. (f) The motion trajectories of all demonstration and automated actions in \(\{C\}\), whose endpoints are aligned.}
    \label{fig:compare}
\end{figure*}

\begin{table*}[t]
    \centering
    \caption{Action Success Rates with Different Methods under Five Evaluation Setups.}
    \label{tab:task_success}
    \renewcommand{\arraystretch}{1.15}
    \begin{tabular*}{\textwidth}{
        @{\extracolsep{\fill}}lccccccc@{}
    }
        \toprule
        \textbf{Method} & \textbf{Standard} & \textbf{Hard Position} & \textbf{New Background}
               & \textbf{New Angle} & \textbf{New FME} & \textbf{Overall} & \textbf{Average Time(s)} \\
        \midrule
        MicroACT-3D \cite{long2025microact}
            & 8/10 & 1/5 & 0/5 & 2/5 & 4/5 & 15/30 & 16.1\\
        HLF \cite{luo2026hierarchical}
            & 9/10 & 2/5 & \textbf{5/5} & 4/5
            & \textbf{5/5} & 25/30 & 16.4\\
        DINOv3ACT
            & 5/10 & 0/5 & 3/5 & 4/5 & 3/5 & 15/30 & 11.7\\
        Ours w/o Adapt.
            & 8/10 & 1/5 & \textbf{5/5} & 3/5 & 3/5 & 20/30 & 12.8\\
        \textbf{Ours}
            & \textbf{10/10} & \textbf{5/5} & \textbf{5/5}
            & \textbf{5/5} & 4/5 & \textbf{29/30} & \textbf{11.5}\\
        \bottomrule
    \end{tabular*}
\end{table*}

\subsection{Implementation Details}

The ROI of the microscopic camera was set to a $384\times384$-pixel square at the image center, corresponding to a field of view of approximately \SI{900}{\micro\meter} in the camera coordinate frame $\{C\}$. The ROI images were fed into a pretrained DINOv3 ViT-B with a patch size of 16 and a stride of 8, and the feature maps from its 6th layer were extracted with a spatial resolution of $47\times47$. End-to-end training was performed using the AdamW optimizer with an initial learning rate of $1\times10^{-5}$, a batch size of 32, and a total of 100 epochs. The number of iterations $L$ in the Object attention module was set to 3. The loss weights $\beta$ and $\gamma$ in Eq. \ref{eq:loss} were both set to 5.
An experienced experimenter collected 65 needle-loop demonstrations using a single needle and a single FME through keyboard teleoperation with real-time monocular microscopic image feedback, taking approximately 30 minutes. During human demonstrations, the initial micro-loop position for each rollout was randomized by adjusting the robot arm while ensuring that the micro-loop remained in focus. The range of position variation was indicated by the green region in Fig.\ref{fig:platform}(b). In addition, the initial microneedle position was randomized for each rollout by adjusting the precision stages. Of the 65 demonstrations, 60 were used to train the visuomotor policy, and the remaining 5 were used for validation. The model was trained on a server equipped with an NVIDIA A40 GPU and subsequently deployed on the robotic platform. The control loop cycle is 150ms and the micro-needle moves in an incremental manner.

\subsection {Comparison Results on Automated Hooking Task}\label{sec:comparison_action}

\textit{1) Evaluation Setups:} To comprehensively evaluate the automated hooking capabilities of visuomotor policies under different conditions, we designed five evaluation setups:

\textit{Standard:} At the beginning of each test rollout, the FME micro-loop position was randomized within the green region in Fig.\ref{fig:platform}(c), matching the range of positions used during human demonstration.

\textit{Hard Position:} At the beginning of each test rollout, the FME micro-loop was placed in one of the purple blocks in Fig.\ref{fig:platform}(c), with the blocks tested in sequence. These blocks were adjacent to the ROI boundary and represented positions that were never used during human demonstration. At these positions, such as Fig.\ref{fig:platform}(d), either a long section of the FME body was visible or only the terminal micro-loop was visible.

\textit{New Background:} As shown in Fig.\ref{fig:platform}(f), this setup differed from the Standard setup in which a piece of blue medical non-woven fabric was placed within the field of view as the background, substantially changing the image contrast and background color.

\textit{New Angle:} As shown in Fig.\ref{fig:platform}(e), this setup differed from the Standard setup in that the FME holder was rotated by approximately $20^\circ$ about the Z-axis of the robot coordinate frame $\{R\}$, introducing a condition that was never encountered during human demonstration.

\textit{New FME:} As shown in Fig.\ref{fig:platform}(g), this setup differed from the Standard setup in that a different FME model was used as the hooking target, with substantial differences in shape, texture, and thickness compared with the FME used during human demonstrations.

 \begin{figure*}[t]
    \centering
    \includegraphics[width=\linewidth]{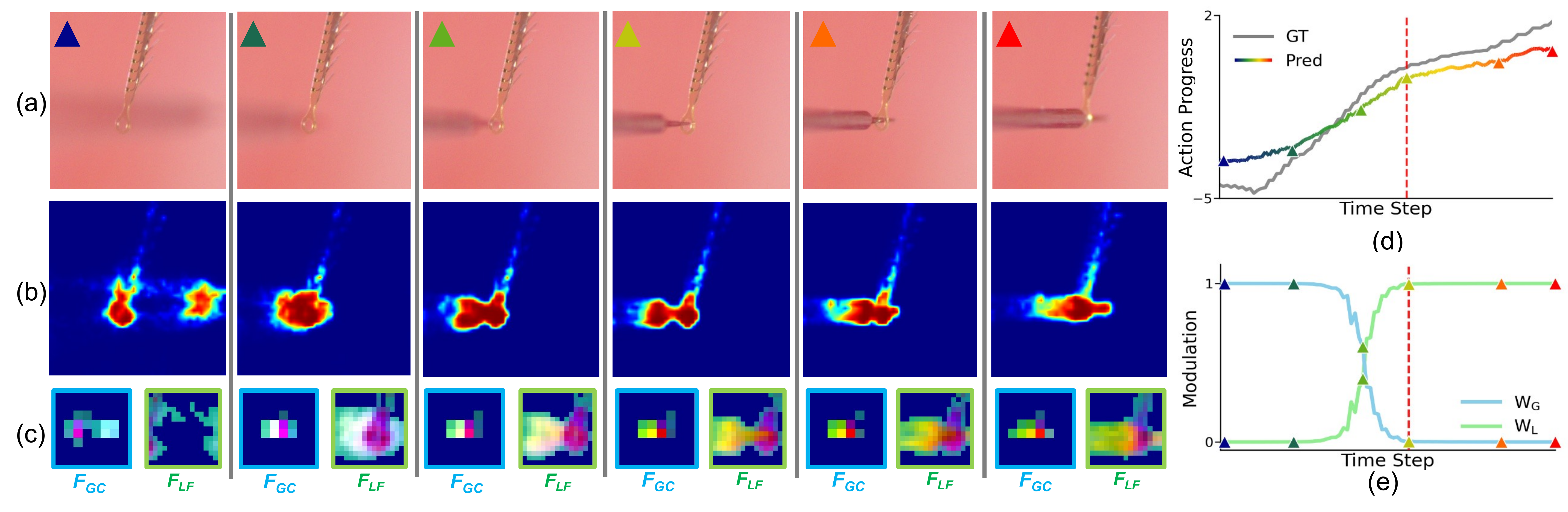}
    \vspace{-25pt}
    \caption{Visualization of object attention, ACGL features, progress prediction, and adaptive modulation. The six columns in (a–c) correspond to the six time steps in (d, e), with matching colored triangles indicating the correspondence.}
    \label{fig:visualize_all}
\end{figure*}
The micro-needle used for evaluation was different from the one used for demonstration. In each rollout, the needle tip position was randomly initialized within the field of view and was initially outside the depth of field, providing sufficient room for adjustment through the unidirectional coarse-to-fine action. Each method was evaluated over 10 rollouts in the Standard setup and 5 rollouts in each of the other setups, for a total of 30 rollouts. A hooking action is considered successful when at least half the length of the micro-needle’s stepped tip has passed through the micro-loop. Premature stopping or movement of the needle tip outside the field of view is considered a failure. 

\textit{2) Comparison Methods:} 
We compared the proposed method with two recent methods, MicroACT \cite{long2025microact} and the hierarchical learning framework (HLF) \cite{luo2026hierarchical}. We made minimal customizations to the original MicroACT by removing the absolute position condition and changing its output from 2D absolute motion to 3D relative motion. In the HLF implementation, we divided the entire hooking action into three stages: coarse alignment, fine alignment, and threading. In addition, we compared our method with ACT using DINOv3 raw features as visual conditioning (DINOv3ACT) and ACT without the action progress based adaptive modulation mechanism (Ours w/o Adapt.) to validate the effectiveness of the proposed object attention and adaptive modulation modules. To ensure a fair comparison, all methods used the same ACT policy architecture and training configuration.

\textit{3) Evaluation Results:} 
Table \ref{tab:task_success} reports the success rates of the five methods. In the Hard Position setup, our method achieved successful hooking at these extreme positions by focusing visual cues on the target and switching from global coarse feature to local fine feature. In comparison, the other four methods exhibited failures in the Hard Position setup, including movements in incorrect directions, premature stopping, and movements outside the field of view.
In the New Background setup, both MicroACT-3D and DINOv3ACT showed substantial performance degradation. Both methods used raw image features as conditioning and modeled actions using a single non-adaptive model, making them sensitive to domain shift. In the New Angle setup, MicroACT-3D and Non-adaptive MicroHookACT showed a relative decline in performance, indicating that changes in the FME angle affected the visual conditioning.
In the New FME setup, HLF achieved the highest success rate, suggesting that decomposing the complete action sequence into three distinct stages simplified the visuomotor coupling within each stage and thereby improved generalization. In the failed rollout of our method, the micro-needle tip grazed the inner edge of the micro-loop, causing the FME to bend with the needle tip and preventing successful threading.
Overall, our MicroHookACT achieved the highest overall success rate, succeeding in 29 of 30 rollouts and demonstrating adaptability to extreme positions, background changes, angle variations, and an unseen FME. HLF also achieved a relatively high success rate. However, it failed to generalize to extreme positions and required a more complex training procedure that relied on manually segmenting the action process and separately training three distinct ACT policies. In addition, we also report the average execution time over successful trials for each method. With a fixed control period, execution time is mainly influenced by trajectory length. As shown in Table \ref{tab:task_success}, our method achieved a comparatively shorter execution time.

\textit{4) Visualized Results:} 
As shown in Fig.\ref{fig:compare}(a), all methods generated correct relative motion directions under the Standard setup. However, the prediction accuracy for coarse approaching motion was substantially affected as the task difficulty increased. For example, in the Hard Position test shown in Fig.\ref{fig:compare}(b), both MicroACT and HLF predicted incorrect motion directions, while Ours w/o Adapt. produced nearly zero motion. In the New Angle test shown in Fig.\ref{fig:compare}(d), the micro-needle was both severely blurred by defocus and close to the boundary of the field of view. Under this condition, both HLF and Ours w/o Adapt. predicted incorrect motion directions that could even move the micro-needle tip outside the field of view.
In addition, we plotted the 60 demonstration trajectories and all 30 MicroHookACT action trajectories in the same coordinate frame, aligning their endpoints to illustrate the motion patterns. The initial micro-needle positions in the evaluation experiments spanned a substantially wider range than those in the demonstration trajectories, demonstrating the strong generalization capability of MicroHookACT to out-of-distribution initial positions.

\subsection{Qualitative Analysis}

\textit{1) Action-Supervised Object Attention:} As shown in Fig.\ref{fig:visualize_all}(b), the proposed action-supervised object attention mechanism kept spatial attention focused on the micro-needle tip and micro-loop throughout the hooking action, while suppressing attention to the micro-needle body, FME body, and background. Even when the micro-needle was severely blurred by defocus, the mechanism localized the region near the micro-needle tip based on its blurred shape.

\textit{2) ACGL Features:} The attention-centered global and local features were visualized in Fig.\ref{fig:visualize_all}(c) via unified PCA-based dimensionality reduction. Global coarse features \(F_{GC}\) encoded the relative positions of the micro-needle tip and micro-loop over a broad spatial range. They therefore exhibited substantial changes during coarse approaching but changed only slightly during fine alignment and threading. Local fine features \(F_{LF}\) primarily contributed when the micro-needle tip and micro-loop overlapped in the XY field of view and were sensitive to micron-level needle-loop interactions. Subtle differences from coarse overlapped to fine aligned and then to threading were reflected in the local fine feature map.

\textit{3) Action Progress Prediction and Adaptive Modulation:} As shown in Fig.\ref{fig:visualize_all}(d), the object slot feature obtained through multiple iterations and global feature aggregation enabled the estimation of the current action progress. The variations in the modulation coefficients calculated using Eq. \ref{eq:routing} were shown in Fig.\ref{fig:visualize_all}(e). The sigmoid function enabled a smooth transition in the relative importance of \(F_{GC}\) and \(F_{LF}\), with action conditioning relying entirely on \(F_{LF}\) in the later stages. Thus, the adaptive modulation module dynamically adjusted the relative weights of the coarse and fine features based on the predicted action progress, gradually shifting the focus of visual guidance from the overall spatial layout to local interaction details.

\subsection{Comparison of Visual Backbones}

We further compared the effectiveness of two visual backbones, ResNet-18 and DINOv3 ViT-B, for action-supervised object attention under both frozen and learnable settings. We randomly selected 50 images with relatively clear needle tips from the automated hooking experiments in Section \ref{sec:comparison_action} and precisely annotated the bounding boxes of the micro-needle tip and micro-loop as the ground-truth object regions. We then thresholded the model-generated object attention maps at 0.5 to obtain the object attention regions and computed two metrics against the ground-truth object regions: \textit{Attention Recall} and \textit{}{Attention Focality}. The former was defined as the overlap area between the attention region and the ground-truth region divided by the ground-truth area, while the latter was defined as the overlap area divided by the attention region area. As shown in Table \ref{tab:ablation}, the frozen DINOv3 ViT backbone presented the significantly superior action focality.

\begin{table}[t]
\caption{Evaluation of Object Attention across Vision Backbones}
\label{tab:ablation}
\centering
\footnotesize
\setlength{\tabcolsep}{2.5pt}
\begin{tabular}{lcccc}
\toprule
\textbf{Backbone} & \textbf{Attention Recall} & \textbf{Attention Focality} \\
\midrule
Resnet-18 (learnable)     & \textbf{0.99}      & 0.21 \\
Resnet-18 (frozen)       & 0.95     & 0.21  \\
DINOv3-ViT (learnable)       & \textbf{0.99}      & 0.02  \\
DINOv3-ViT (frozen)         & 0.97       & \textbf{0.35}  \\
\bottomrule
\end{tabular}
\end{table}

\section{CONCLUSIONS}
This work presented MicroHookACT for automated FME hooking under monocular microscopic vision. By combining a unidirectional hooking strategy with action-supervised object attention and progress-based adaptive modulation of global coarse and local fine features, the framework enabled a single visuomotor policy to perform coarse approaching, fine alignment, and contact-rich threading without palpation or separate stage-specific policies. Future work will study the transferability of visuomotor policy across different types of needle and FME.

\bibliographystyle{IEEEtran}   % 如果文档类是 IEEEtran，用这个；否则可用 ieeetr
\bibliography{reference}       % 不要写 .bib 后缀

\end{document}